\documentclass[runningheads,a4paper]{llncs}

\usepackage{graphicx}

\usepackage{hyperref}
\usepackage{tabu, booktabs}

\usepackage{algorithm, algorithmic}

\usepackage{amsmath, amsfonts, amsthm}

\usepackage[utf8]{inputenc}

\newcommand{\goth}[1]{\ensuremath{\mathcal{#1}}}
\newcommand{\normal}{\goth{N}}
\newcommand{\Cov}{\mathbf{\Sigma}}
\newcommand{\abs}[1]{\lvert#1\rvert}

\newcommand{\keywords}[1]{\par\addvspace\baselineskip
\noindent\keywordname\enspace\ignorespaces#1}

\begin{document}

\mainmatter 

\title{Towards Gaussian Bayesian Network {Fusion}}

\titlerunning{Towards Gaussian Bayesian Network Fusion}

\author{Irene Córdoba-Sánchez\and Concha Bielza\and Pedro Larrañaga}

\authorrunning{I. Córdoba-Sánchez, C. Bielza, P. Larrañaga}

\institute{Departamento de Inteligencia Artificial\\
Universidad Politécnica de Madrid\\
\path{irene.cordoba.sanchez@alumnos.upm.es}\\
\path{{mcbielza, pedro.larranaga}@fi.upm.es}}

\toctitle{Towards Gaussian Bayesian Network Fusion}
\tocauthor{Irene Córdoba-Sánchez, Concha Bielza, Pedro Larrañaga}
\maketitle

\begin{abstract}
Data sets are growing in complexity {thanks to the increasing facilities we have nowadays to both generate and
store data. This poses many challenges to machine learning} that are leading to the proposal of new methods and
paradigms{, in order to be able to deal with what is nowadays referred to as Big Data}.
In this paper we propose a {method for the aggregation of different Bayesian network structures that 
have been learned from separate data sets, as a first step towards mining} data sets that {need to be partitioned
in an horizontal way, i.e. with respect to the instances, in order to be processed}. 
Considerations that should be taken into account when dealing with this {situation} are discussed. Scalable learning
of Bayesian networks is slowly emerging, and our method constitutes one of the first insights into Gaussian Bayesian
network aggregation from different sources.
Tested on synthetic data it obtains good results {that surpass those from individual learning}. Future research will
be focused on {expanding the method and testing more diverse data sets}.
\end{abstract}

\section{Introduction}

Nowadays, we are entering the \textit{era of Big Data}, as a result of 
both the generalised trend of massive data collection and the increasing computer capabilities for processing and storage. 
{These data sets are characterized mainly for their huge volume and complexity (they can be noisy, have a fast change rate, etc.).
Machine learning methods are rapidly being revised and new paradigms are arising in order to be able to adapt to this kind of
data. 

One of the main approaches for dealing with high volume of data is to partition it across a 
cluster, perform some operations and then aggregate the results. This partition can be either horizontal (across the instances) or
vertical (across the variables). Horizontal partitions can also naturally arise when we want to jointly analyse information
contained at different sources, e.g. records of patients in different hospitals that store the same variables about each of them.}

{Bayesian networks (BNs) are well-known tools for modelling and dealing with uncertain knowledge and data. 
Their aggregation has been studied since the days of their conception as belief models from
an expert. Martzkevich and Abramson \cite{matzkevich1992} consider the problem of fusing
networks from different experts which shared some variables. They provide an algorithm which seeks to obtain a graph containing all
the nodes and arcs from the individual networks, or their reversals. This however may not be the case of interest always when we
are thinking about fusing networks that have been learned from data, since the individual networks in this case may contain spurious 
connections.}

{The work by del Sagrado and Moral \cite{sagrado2003} focuses on studying the fusion of DAGs by means of intersection
and union of the independence statements represented by each of the involved networks.
Richardson and Domingos \cite{richardson2003} use knowledge from a group of experts to compute a prior distribution over the BN structures. They motivate their proposal by stating that knowledge elicitation can be 
facilitated if we allow experts to be noisy on their statements about the BNs, and make up for this flexibility by using
multiple different experts. This argument is interesting because it can be compared with the case of huge, noisy data sets,
where instead of sub-sampling and learning an individual network, an alternative approach could be to learn multiple
networks on different partitions of data and aggregate them afterwards.

Another use case where horizontal partitioning arises naturally is the problem described by López-Cruz \emph{et al.} \cite{lopezcruz2014}. 
In this case a set of experts were asked to classify different neurons, giving rise to one supervised training set from each expert.
A cluster process was applied to the set of the individual BNs obtained and a representative BN for each
cluster was constructed. These representative networks were then aggregated into a Bayesian multinet.}

{{Regarding the aggregation of parameters, recently Etminani \emph{et al.} \cite{etminani2013} propose a method in which they cluster experts' parameters and aggregate only those that correspond to the cluster with the highest number of members, resembling democratic societies. Other popular strategies for parameter fusion in Bayesian networks are Linear Opinion Pools (LinOP) \cite{maynard2001} and Logarithmic Opinion Pools (LogOP) \cite{pennock1999}.}}

We propose a method for {the aggregation of} Gaussian BNs (GBNs), which to the best of our 
knowledge is the first proposal {of this kind. It covers both the structure of the network
and the parameters of the Gaussian distribution encoded by it. 
The experiments carried out show promising results for the proposed method.}

The paper is organised as follows. Section \ref{sec:prel} introduces the necessary background knowledge for
the rest of the paper. In Section
\ref{sec:met} the details of the method are described, whose results from experimental evaluation
are discussed in Section \ref{sec:exp}. Finally, the conclusions and
future research lines are presented in Section \ref{sec:con}. 

\section{Preliminaries} \label{sec:prel}

\subsection{Bayesian networks}

A BN can be defined as a way of representing the factorization of a joint probability distribution over a random vector $\boldsymbol{X} = (X_1, ..., X_p)$,
where $\boldsymbol{Pa}(X_i)$ are called the \textit{parents} of $X_i$,
\begin{equation}\label{eq:fact}
	f(\boldsymbol{x}) = \prod_{i=1}^p f(x_i| \boldsymbol{pa}(x_i)).
\end{equation}
A BN consists on a qualitative part, commonly called the structure, and a quantitative component, the parameters. 
More formally, it is defined \cite{pearl1988} as a pair $(G, \Theta)$, where $G$ is a DAG
and $\Theta$ are the numerical parameters which define the factorization in Equation (\ref{eq:fact}). 
The nodes of $G$ are the components of $\boldsymbol{X}$ and its arcs represent 
probabilistic dependencies between the variables, in such a way that the DAG satisfies the \textit{Markov condition}: each variable is conditionally independent of its non-descendants given its parents. 
Two DAGs are \textit{Markov equivalent} if they represent the same set of conditional independences between the variables. This
defines a binary relation which gives rise to equivalence classes and partitions the DAG space.

In order to learn a BN from data it is necessary to learn both the structure ($G$) and the numerical parameters ($\Theta$). 
There are two main approaches for BN structure learning: constraint based and score-and-search.
Constraint based methods try to find the Bayesian network structure that represents most of the dependence relations present
in data, detected by means of statistical tests. The PC algorithm \cite{spirtes2000}, which has as output an equivalence
class of DAGs, is a representative example of these types of methods.

On the other hand, score-and-search methods try to find the structure that best fits the data. They are characterized by
a representation of the solution space, a search method and a score. The KES algorithm \cite{nielsen2002} is an example of such methods, 
which performs the search in the equivalence class space. Searching in this space has several
advantages when compared to the DAG space, such as it being a more efficient and robust representation \cite{vidaurre2010},
although there is still some controversy regarding this choice. Many search heuristics and scores can be combined
and give rise to the different methods appearing in the literature.

\subsection{Gaussian Bayesian networks}

A GBN \cite{geiger1994} encodes a joint Gaussian distribution over $\boldsymbol{X}$, i.e., with joint density function
$$f(\boldsymbol{x}) = \frac{1}{\sqrt{(2\pi)^p\abs{\Cov}}}\exp\left\{-\frac{1}{2}(\boldsymbol{x}-\boldsymbol{\mu})^t\Cov^{-1}(\boldsymbol{x}-\boldsymbol{\mu})\right\},$$
where $\boldsymbol{\mu} = (\mu_1, ..., \mu_p)$ is the vector of unconditional means and $\Cov$ is the covariance matrix. 
Each factor in Equation (\ref{eq:fact}) corresponds in this case to a univariate normal distribution,
\begin{equation}\label{eq:gauss}
	f(x_i| \boldsymbol{pa}(x_i)) \equiv \normal\left(\mu_i + \sum_{x_j \in \boldsymbol{pa}(x_i)}\beta_{ji}(x_j - \mu_j), v_i\right),
\end{equation}
where $\beta_{ji}$ reflects the strength of the relationship between $X_i$ and its $j$-th parent, and
$v_i$ is the conditional variance of $X_i$ given its parents, i.e.,
\begin{equation} \label{eq:condv}
v_i = \sigma_i - \Cov_{i\boldsymbol{Pa}(X_i)}\Cov_{\boldsymbol{Pa}(X_i)}\Cov_{i\boldsymbol{Pa}(X_i)}^t.
\end{equation}
In Equation \ref{eq:condv} $\sigma_i$ is the unconditional variance of $X_i$, $\Cov_{i\boldsymbol{Pa}(X_i)}$ is the matrix of covariances
between $X_i$ and $\boldsymbol{Pa}(X_i)$, and $\Cov_{\boldsymbol{Pa}(X_i)}$ is the covariance matrix of $\boldsymbol{Pa}(X_i)$.

Thus, the parameters of a GBN are the vector of means $\boldsymbol{\mu}$, the vector of conditional
variances $\boldsymbol{v}$ and the coefficients $\beta_{ji}$.
Assuming standardized data ($\mu_i = 0$ and $v_i = 1$), the parameter estimation is reduced to solving the linear regression model
$$x_i = \sum_{x_j \in \boldsymbol{pa}(x_i)}\beta_{ji}x_j + \epsilon_i,$$
with $\epsilon_i$ being the Gaussian noise term with zero expectation.

\section{Method} \label{sec:met}

{Although the aggregation of different individual GBNs is a first
step towards the analysis of massive data, where} the data set {would be} split into slices distributed across
{a cluster, here we will assume that
we already have different data sets over the same variables available (i.e., at this stage we are not concerned with the
preprocessing and splitting processes)}. 

The structure learning method we have used for learning the individual networks is the score-and-search hill climbing \cite{tsamardinos2006} with the Bayesian information criterion (BIC) \cite{schwarz1978} score on the DAG space. {After each network has been learned, they are aggregated
using majority vote below a threshold. This procedure is outlined in Algorithm \ref{alg:struc}}. 

\begin{algorithm}
{
\caption{Structure learning}\label{alg:struc}
\begin{algorithmic}[1]
	\REQUIRE \emph{datasets}. Data sets from where the individual Bayesian networks will be learned.
	\REQUIRE \emph{threshold}. Threshold for the majority arc voting.
	\ENSURE Aggregated Bayesian network structure learned with the specified thresholds.
	\STATE $n\_bn \gets $size$(datasets);$
	\STATE $bn\_list \gets $list$();$
	\FOR{$i \in \{1, n\_bn\}$}
		\STATE $bn\_list[i] \gets $learn\_struc$(datasets[i])$
	\ENDFOR
	\STATE $v\_matrix \gets$ get\_votes($bn\_list$);
	\STATE $result \gets $bn\_aggr$(threshold, v\_matrix);$
	\RETURN $result;$
\end{algorithmic}
}
\end{algorithm}

{The functions $get\_votes$ and $bn\_aggr$ in Algorithm \ref{alg:struc} are further detailed in Algorithms \ref{alg:votes} and \ref{alg:aggr} respectively. 
$get\_votes$ 
consists of the process of extracting how many networks contribute to the same arc, i.e., how common across the 
learned networks an arc is. Thus a matrix containing the sums of the appearances of each arc in the networks is obtained.
The threshold for the majority vote is the main parameter of this method and we will analyse it further on the experimental
section.}


%

\begin{algorithm}
{
\caption{get\_votes}\label{alg:votes}
\begin{algorithmic}[1]
	\REQUIRE \emph{bn\_list}. List of BN structures already learned on each data set.
	\ENSURE Matrix containing the votes for each arc.
	\STATE $n\_nodes \gets $nodes$(bn);$
	\STATE $v\_matrix \gets $matrix$(n\_nodes, n\_nodes);$
	\FOR{$bn \in bn\_list$}
		\STATE $bn\_arcs \gets $arcs$(bn);$
		\FOR{$arc \in bn\_arcs$}	
			\STATE $from \gets $from$(arc);$
			\STATE $to \gets $to$(arc);$
			\STATE $v\_matrix[from][to] \gets v\_matrix[from][to] + 1;$
		\ENDFOR
	\ENDFOR
	\RETURN $v\_matrix;$
\end{algorithmic}
}
\end{algorithm}

{In $bn\_aggr$ the threshold is compared with each of the entries in the matrix of arcs, and the corresponding arc is
added to the final network if its value reaches the threshold. In the same algorithm we can notice that} when an arc 
addition causes a cycle in the DAG it is discarded. 

\begin{algorithm}
{
\caption{bn\_aggr}\label{alg:aggr}
\begin{algorithmic}[1]
	\REQUIRE \emph{threshold}. Threshold for the arc voting.
	\REQUIRE Matrix containing the votes for each arc.
	\ENSURE \emph{bn}. Aggregated Bayesian network.
	\STATE $bn \gets$ empty\_dag();
	\FOR{$i \in $ cols$(v\_matrix)$}
		\FOR{$j \in $ rows$(v\_matrix)$}	
		\IF{$v\_matrix[i][j] \ge threshold$}
			\IF{\NOT arc\_causes\_cycle$(bn, i, j)$}
				\STATE add\_arc$(bn, i, j);$
			\ENDIF
		\ENDIF
		\ENDFOR
	\ENDFOR
	\RETURN $bn$;
\end{algorithmic}
}
\end{algorithm}

After the {aggregation of the structure} has finished, the linear regression coefficients
of each variable on its parents {is learned} by maximum likelihood estimation (MLE) {from each
data set, but this time using the aggregated structure. This is what del Sagrado and Moral \cite{sagrado2003} call
\emph{topological fusion}, that is, obtaining a  consensus structure and then estimating the model parameters,
as opposed to \emph{graphical representation of consensus}, which consists of aggregating the probability distributions
of each network and then obtaining the structure that represents it.}


The {aggregation} of the parameters obtained {from each data set is performed} using the method explained hereafter.
Consider a multiple linear regression model on $\{X_1, ..., X_n\}$ predictors. Assume that the data is {distributed across}
$k$ slices. Let $\hat{\boldsymbol{\beta}}_j = (\hat{\beta}_{1j}, ..., \hat{\beta}_{nj})$ be the vector of 
estimates obtained in slice $j$. For each predictor $X_i$, $i \in \{1, ..., n\}$, let 
	$$\displaystyle{\tilde{\beta_i} = \sum_{j=1}^k w_{ij}\hat{\beta}_{ij}}$$
be the aggregated estimate, where $w_{ij} = \sigma_{ij}^{-2}/\sum_{j=1}^k \sigma_{ij}^{-2}$, $\sigma_{ij} = var(\hat{\beta}_{ij})$.
Because we are dealing with GBNs, MLE is equivalent to the least squares (LS) method, and thus
$\dot{\beta}_i$ is the estimator of minimum variance \cite{fan2007} among those with form 
$$\sum_{j=1}^k w_{ij}\hat{\beta}_{ij}, \text{ where }  \sum_{j=1}^k w_{ij} = 1,$$ 
Asymptotic normality is also established on Fan \emph{et al.} \cite{fan2007}.

The pseudo-code {of the outlined procedure for learning the parameters of the linear regression
for each variable on its parents can be found in} Algorithm \ref{alg:param}.

\begin{algorithm}
\caption{Parameter learning}\label{alg:param}
\begin{algorithmic}[1]
	\REQUIRE {\emph{datasets}. Data sets from where the individual parameters will be learned.}
	\ENSURE Bayesian network parameters {aggregated}.
	\STATE $n\_bn \gets $size$(datasets);$
	\STATE $param\_list \gets $list$();$
	\FOR{$i \in \{1, n\_bn\}$}
		\STATE $param\_list[i] \gets $learn\_param$(datasets[i]);$
	\ENDFOR
	{
	\STATE $n\_param \gets $size$(param\_list);$
	\STATE $n\_nodes \gets $nodes$(param\_list);$
	\STATE $param \gets $matrix$(n\_nodes, n\_nodes);$
	\FOR{$i \in \{1, n\_param\}$}
		\FOR{$node \in param\_list[i]$}
			\FOR{$parent \in $ parents$(node)$}
				\STATE $coef \gets $get\_coeff($param\_list[i], node, parent$);
				\STATE $weight \gets $get\_weight($param\_list[i], node, parent$);
				\STATE $param[node][parent] \gets param[node][parent] + coef*weight;$
			\ENDFOR
		\ENDFOR
	\ENDFOR
	\RETURN normalize($param$);
	}
\end{algorithmic}
\end{algorithm}

\section{Experiments} \label{sec:exp}

\begin{figure}
	\centering
	\includegraphics[width=0.4\linewidth]{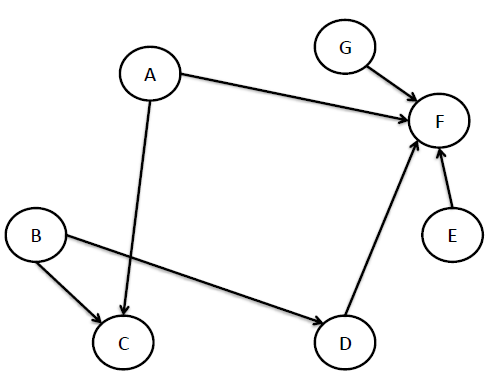}
	\caption{Structure of the Bayesian network used {for the experiments}.}
	\label{fig:bntest}
\end{figure}

We have used some utilities from 
the R package {\tt bnlearn} \cite{scutari2010} {and tested the 
proposed method using synthetic data sets generated} from a multivariate Gaussian distribution
whose DAG structure is shown in Figure \ref{fig:bntest}.

{In a real use case of this method we could have been given a number of
separate data sets over the same variables but differing in the number of instances each one contains. On the other hand, if we were to apply it to a huge data set, the different partitions would probably contain a similar amount of instances. In this synthetic experiment we have generated eight different data sets with a sample size of 50 instances each.}

\begin{figure}
	\centering
	\includegraphics[width=\linewidth]{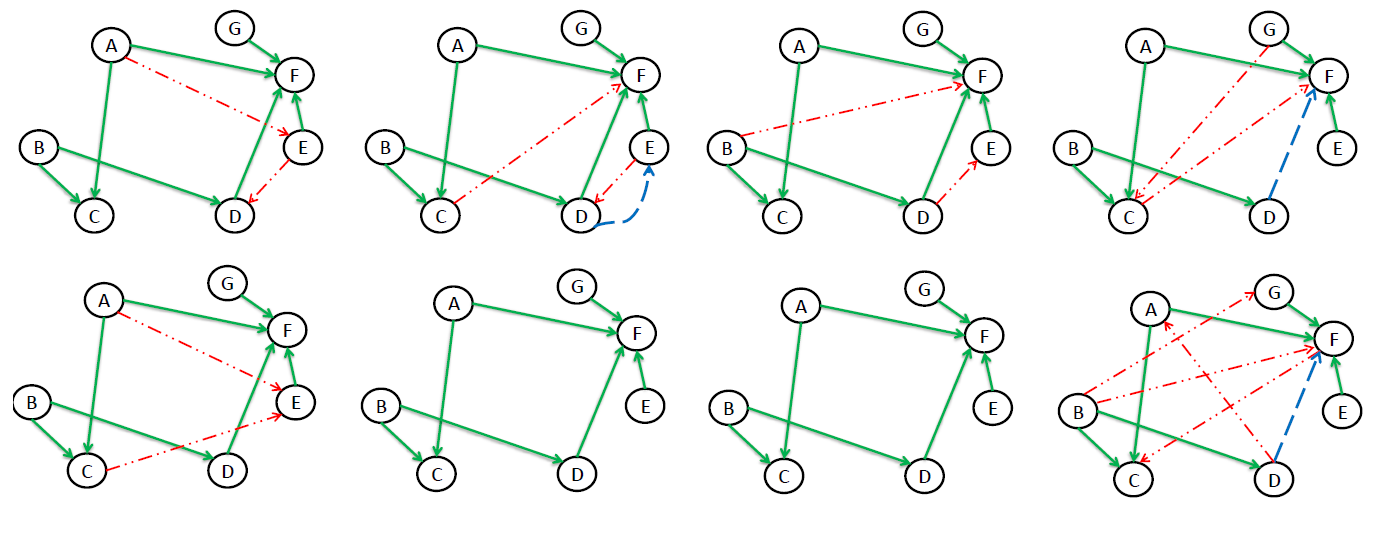}
	\caption{{Structure learned on each of the data sets. Green arcs are those correctly learned, red arcs are false positives and blue arcs are false negatives.}}
	\label{fig:bnresults}
\end{figure}

\begin{figure}
	\centering
	\includegraphics[width=\linewidth]{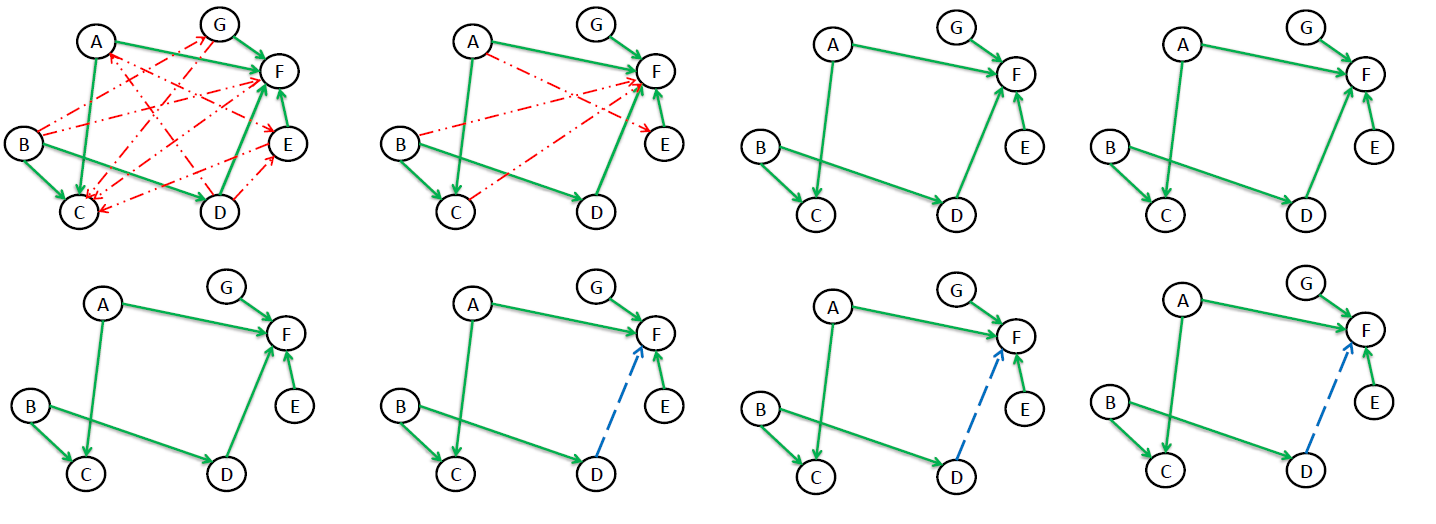}
	\caption{{Aggregated structures for the different thresholds (1 to 8), increasing from left to right and top to bottom. Green arcs are those correctly aggregated, red arcs are false positives and blue arcs are false negatives.}}
	\label{fig:merge}
\end{figure}

{Figure \ref{fig:bnresults} shows the different BN structures obtained from each data set. We can notice that a high portion of the original network is learned in most of the cases, being false positives the most common error.}
{We have aggregated the results using all possible values for the threshold parameter, getting as result the networks that appear on Figure \ref{fig:merge}.}

{The metrics we are going to use to evaluate the results obtained with respect to the original structure are the false positive, false negative and true positive rate and the Structural Hamming Distance (SHD) \cite{tsamardinos2006}. The latter consists of the number of operations needed to match the Partial DAGs (PDAGs) representing the equivalence classes of each network. The operations considered are arc addition, deletion or reversal and edge addition or deletion. Therefore, the PDAG is extracted from the respective DAGs before calculating this metric. SHD provides a way to compare the two BNs in terms of the conditional independencies encoded by the BN, and thus avoids the penalization of differences in arcs that might be statistically undistinguishable.}

{In Table \ref{tab:sresults} we can see the value metrics for the individual structures (left) and the aggregated ones (right) when compared with the original network.}
\begin{table}
{
\centering
		\begin{minipage}[t]{0.45\textwidth}
		\begin{tabular} {c c c c c }
		\toprule
		Network & SHD & TP & FP & FN\\
		\midrule
		1 & 3 & 7 & 2 & 0\\
		2 & 3 & 6 & 2 & 1\\
		3 & 2 & 7 & 2 & 0\\
		4 & 3 & 6 & 2 & 1\\
		5 & 2 & 7 & 2 & 0\\
		6 & 0 & 7 & 0 & 0\\
		7 & 0 & 7 & 0 & 0\\
		8 & 5 & 6 & 4 & 1\\
		\bottomrule
		\end{tabular}
		\end{minipage}
		\begin{minipage}[t]{0.45\textwidth}
		\begin{tabular} { c c c c c }
		\toprule
		Threshold & SHD & TP & FP & FN \\
		\midrule
		1 & 8 & 7 & 8 & 0 \\
		2 & 3 & 7 & 3 & 0 \\
		3 & 0 & 7 & 0 & 0 \\
		4 & 0 & 7 & 0 & 0 \\
		5 & 0 & 7 & 0 & 0 \\
		6 & 1 & 6 & 0 & 1 \\
		7 & 1 & 6 & 0 & 1 \\
		8 & 1 & 6 & 0 & 1 \\
		\bottomrule
		\end{tabular}
		\end{minipage}
	\caption{{Results of the GBN learned on each data set (left) and the aggregated GBN (right) compared with the original network}. TP, FP and FN indicate the true positive, false positive and false negative rates (respectively). SHD denotes the Structural Hamming Distance. The networks are numbered according to their order of appearance in Figure \ref{fig:bnresults}}
	\label{tab:sresults}
	}
\end{table}

{Obviously when the threshold is 1 every arc that appears in the individual networks is added to the final one (unless a cycle is caused), so this produces worse results than any individual network when comparing it to the original structure. However, for thresholds above 1 the aggregated result is better than most of the isolated ones, because thanks to the majority threshold false dependences are eliminated. For too restrictive thresholds this can however result in the deletion of a valid arc, so it should be adjusted to an intermediate value for the best results. In a real use case this would depend on the data characteristics, the application domain and the availability of a training set.}

{Finally, parameter learning is influenced by noise if the structure is not correctly learned because false parents of variables arise, which means that false coefficients are estimated in the linear regression. However it is the case, as one would expect, that the coefficients corresponding to these false parents are very close to zero (e.g., 0.007 mean for the extreme case of threshold 1) and the
variations on the value of the other parents are barely noticeable. This is not the case when we learn from a single network. For 
example, in the last network in Figure \ref{fig:bnresults}, $\beta_{BF} = 1.5$, and $B$ is a false parent of $F$.}

		%

\section{Conclusions and Future Work} \label{sec:con}
{We have considered the problem of horizontal partitioning in the context of Big Data and} proposed a 
method for {aggregating several} GBNs {learned from different data sets as a first step
towards} scalable GBN learning. The method obtains
good results both in the case of structure and parameter learning on synthetic data. {The aggregated results
surpass in most cases those derived from learning from a single data set by taking into account all the data
available without the need of analysing it as a single block. This is specially useful for its potential applications
when analysing partitions of massive data sets.}

{{As a future line of research,} when learning the aggregated structure the
treatment of cycles will be refined and will involve more sophisticated techniques such as arc reversal, checking the strength of the connection in each
of the individual networks (coefficients of the regression), establishing a suitable ordering of arc consideration, etc.}

{In the case of applying the proposed method on a distributed setting, where each data set is in a computer within a cluster, it would be interesting to define some communication protocol during the learning process. This would be useful for gathering stepwise information that could be used later on for example when aggregating the individual networks (e.g. cycles) but also for developing more sophisticated voting schemes which could depend on the adequateness of each data set for the learning process.}

{Finally, } we will {also} focus on {performing more testing with diverse (noise, missing values) and real data sets.}

\section*{Acknowledgements} \label{sec:ack}
{The authors thank the reviewers for comments and critics
which significantly contributed to improve the paper; and also} J.M. Peña, J. Nielsen{, J. Mengin and M. Serrurier} for the valuable help. This work has been partially supported by the Spanish Ministry of Economy and Competitiveness through the Cajal Blue Brain (C080020-09; the Spanish partner of the Blue Brain initiative from EPFL) and TIN2013-41592-P projects, and by the Regional Government of Madrid through the S2013/ICE-2845-CASI-CAM-CM project.

\end{document}